\documentclass{article}

\usepackage{arxiv}

\usepackage[T1]{fontenc}    
\usepackage{hyperref}       
\usepackage{url}            
\usepackage{booktabs}       
\usepackage{amsfonts}       
\usepackage{nicefrac}       
\usepackage{microtype}      
\usepackage{lipsum}
\usepackage{graphicx}
\usepackage{amsmath}
\usepackage{caption}
\usepackage{subcaption}
\usepackage{hyperref}
\usepackage{caption}
\usepackage{subcaption}
\usepackage{MnSymbol}
\usepackage{wasysym}
\usepackage{graphicx,color,framed,a4wide}
\usepackage{ifpdf}
\usepackage{array}
\usepackage{arabtex}
\usepackage{utf8}

\usepackage{booktabs}
\usepackage[toc,page]{appendix}
\usepackage[math]{cellspace}
\graphicspath{ {./images/} }

\title{Calliar: An Online Handwritten Dataset for Arabic Calligraphy}

\author{
  Zaid Alyafeai  \\
   Department of Computer Science\\
  King Fahd University of Petroleum and Minerals\\
  Dhahran, Saudi Arabia 31261 \\
  \texttt{g201080740@kfupm.edu.sa} \\
     \And
    Maged S. Al-shaibani \\
  Department of Computer Science\\
  King Fahd University of Petroleum and Minerals\\
  Dhahran, Saudi Arabia 31261 \\
  \texttt{g201381710@kfupm.edu.sa} \\
   \And
   Mustafa Ghaleb \\
Department of Computer Science\\
  King Fahd University of Petroleum and Minerals\\
  Dhahran, Saudi Arabia 31261 \\
  \texttt{g200905270@kfupm.edu.sa} \\
  \And
   Yousif Ahmed Al-Wajih  \\
  Department of Systems Engineering\\
  King Fahd University of Petroleum and Minerals\\
  Dhahran, Saudi Arabia 31261 \\
  \texttt{g201472320@kfupm.edu.sa} \\
  
}

\begin{document}
\setcode{utf8}
\maketitle
\begin{abstract}

Calligraphy is an essential part of the Arabic heritage and culture. It has been used in the past for the decoration of houses and mosques. Usually, such calligraphy is designed manually by experts with aesthetic insights. In the past few years, there has been a considerable effort to digitize such type of art by either taking a photo of decorated buildings or drawing them using digital devices. The latter is considered an online form where the drawing is tracked by recording the apparatus movement, an electronic pen for instance, on a screen. In the literature, there are many offline datasets collected with a diversity of Arabic styles for calligraphy. However, there is no available online dataset for Arabic calligraphy. In this paper, we illustrate our approach for the collection and annotation of an online dataset for Arabic calligraphy called Calliar that consists of 2,500 sentences. Calliar is annotated for stroke, character, word and sentence level prediction. 
\end{abstract}

\section{Introduction}
Handwriting is used to convey information by writing on a paper or digital screen by hand. Every person, while writing the same statement, has a unique way of drawing letters. Compare that to digital typing which is mostly invariant and usually conducted using a set  of predefined fonts. Processing handwriting is a difficult task because of its complexity and the different styles of people's writings. Add to that, if we consider a complex language like Arabic which is cursive i.e letters are connected together which causes different variations in the writing system. For example, the letter \<س> is written like that when comes at the end of words, when it comes at the middle it is written like this \<ـسـ> . Similarly, the letter \<ع> is written as \<ـعـ> which can be considered as a completely different letter. The complexity is also even higher when we add diacritics which are used as an alternative for short vowels in Arabic. 

Arabic calligraphy has a long history since it was used for a long time to decorate houses, mosques and public places. Opposed to regular handwriting, calligraphy adds some kind of artistic view owing to the freedom of drawing letters. There are various of calligraphic styles and each one of them has specific styles and rules. The most popular calligraphy types are Diwani \<ديواني>, Thuluth \<ثلث> , Kufi \<كوفي> , Farisi \<فارسي> , Naskh \<نسخ> and Rekaa \<رقعة> . We can consider regular handwriting as a special case of calligraphy where a specific style is used, more prominently: Naskh \<نسخ> and Rekaa \<رقعة>. Another complexity of writing in calligraphy is the freedom of drawing in a convoluted and non-linear approach; which might cause people difficulty in understanding some calligraphic styles. In addition to that, some writers add some special characters like diacritics and other decoration symbols. 

In the literature, there are many studies that collect large datasets for handwriting  in Arabic whether offline or online. Offline writing deals with handwritten text as an image where we don't have any idea how each letter is drawn. On the other hand, online writing records the information of each point and how to draw each single letter i.e each written character can be redrawn latter in an animated way. The most popular datasets are KHATT \cite{mahmoud2014khatt} which consists of offline handwritten text. An example of online dataset is ADAB \cite{margner2009icdar} which composes of 32,492 online pen traces. 

The existing datasets deal only with regular handwriting (as also indicated in the literature review section). To the best of our knowledge, there is no online dataset for Arabic calligraphy. Our main contribution is creating a large online dataset called Calliar for Arabic calligraphy in different styles. The main outcome is a dataset that allows users to extract stroke, character, word and sentence information in an easy manner. Moreover, the dataset consists of different variations of Arabic calligraphy styles like Diwani, Thuluth, Kufi and Farisi. This paper is organized as follows. \textbf{Section 2} reviews the current studies and summarizes the existing Arabic datasets. \textbf{Section 3} and \textbf{4} describe the annotation, visualization and give a summary of our dataset. \textbf{Section 5} and \textbf{6} emphasize the dataset importance by summarizing its impact on research, history and society. 

\section{Related Work}

In the literature, there are many studies that dealt with collecting a large dataset for regular handwritten. One of the largest datasets for English that is used heavily in the literature is the IAM dataset \cite{marti2002iam} which is an offline dataset written by 400 different writers. The total number of word instances in the dataset is 82,227 with a total of 10,841 instances of unique vocabulary dataset.The Casia \cite{liu2011casia} (Chinese handwriting databases) dataset consists of 3.9 million samples of 7,356 classes which is a collection of online and offline datasets.  UPTI \cite{sabbour2013segmentation} dataset consists of more than 10,000 synthetic images of Urdu text written in Nastaleeq font. 

In the field of Arabic handwriting recognition systems, there are many datasets that are offline like KHATT \cite{mahmoud2014khatt}, which consists of 1,000 handwritten forms written by 1,000 different writers from different countries. KHATT was extended later to online-khatt that consists of 10,040 lines conducted by 623 different writers \cite{mahmoud2018online}.  ADAB \cite{margner2009icdar} on the other hand is an online dataset that consists of 32,492 Arabic words handwritten by more than 1000 writers. There are also multilingual datasets that combine Arabic and English like MAYASTROUN \cite{njah2012mayastroun}. It consists of 67,825 samples that were written by 355 writers. The dataset consists of many variant scripts like words, characters, digits, mathematical expressions, and signatures.  

There are a few studies in the literature that deal with Arabic calligraphy.  Koudja et al. \cite{kaoudja2019efficient} collected 1,685 images and categorized them into 9 classes of different calligraphic styles like Thuluth, Naskh, Diwani, etc. Allaf and Al-Hmouz \cite{allaf2016automatic} designed a system for recognizing artistic Arabic calligraphy types. Then, they evaluated the system on three different Arabic handwritten calligraphy styles which are Thuluth, Reqaa, and Kufi. The total size of the dataset is 267 images with equal number of images in each class. There are also challenging datasets that consider historical texts like KERTAS \cite{adam2018kertas} which composes of around 2,000 images taken from various handwritten Arabic scripts mainly during the fourteen centuries. AlSalamah et al. \cite{salamah2018towards}  collected 1,000 calligraphy images that were scrapped from public websites. The dataset was manually annotated to extract letter images. Bataineh et al. \cite{bataineh2013arabic} collected 700 samples of Diwani, Rekaa, Kufi, Persian, Andalusi, Naskh, and Thuluth consisting of 100 images, gathered from several sources. In another research, Bataineh et al. \cite{bataineh2011arabic} collected a dataset that consists of 14 Arabic degraded document images. Each image represents one of binarization challenges  such as law contract, seeking ink, multi-color and dirty spots. Allaf et al.  \cite{allaf2016automatic} developed an algorithm for Arabic calligraphy types recognition. Then, they tested their algorithm on a combination of local and public dataset. The local dataset consists of 18 images of sentences written by a skilled calligraphist. In their work, words are separated manually from the sentence images and some of the words are rewritten independently since they could not split it from the original sentence. The total number of samples in this dataset is 71 words. There is a also big computer-generated dataset called APTI (Arabic Printed Text Image) \cite{slimane2009new}. This dataset was generated utilizing lexicon of 113,284 words with 20  different fonts and sizes in 4 different calligraphy styles. Khayyat et al. \cite{khayyat2020deep} collected 2,653 images of a historical Arabic manuscripts. The dataset is categorized into 37 classes with six handwriting styles. Kaoudja at al. \cite{kaoudja2019efficient} proposed a dataset consisting of 9 calligraphy styles. In total,  1,685 images have been classified under their inscribed styles. Each calligraphy style consists of around 180 to 195 images. Adam et al. \cite{adam2017based} collected 330 images of isolated Arabic letters that were extracted from ancient manuscripts. This dataset consists of Rekaa, Diwani, Kufi, Naskh, and Farsi styles. This dataset has been used to classify Arabic scripts style based on segmented letters. 

As apparent from the aforementioned studies and up to our knowledge, there is no dataset in the literature that deals with Arabic calligraphy in online format.  In Table \ref{tab:lit}, we summarize the existing datasets based on four criteria named number of samples, type, language and either the dataset is public or private. 
\begin{table}[htp!]
\centering
\begin{tabular}{l|l|l|l|l}
\textbf{Reference} & \textbf{Number of samples}      & \textbf{Type}                                          & \textbf{Language} & \textbf{Public} \\ \hline \hline
Allaf et al. \cite{allaf2016automatic}    &  18               & Sentences                    & Arabic   &         \\ \hline
Al-Hmouz \cite{allaf2016automatic}        & 267              & Sentences                  &  Arabic   &         \\ \hline
Adam et al. \cite{adam2017based}           & 330              & Characters                   &  Arabic  &             \\ \hline
KHATT \cite{mahmoud2014khatt}   & 1,000             & Sentences              & Arabic     & $\checkmark$       \\ \hline
online-khatt \cite{mahmoud2018online}   & 10,040              & Sentences              & Arabic     & $\checkmark$       \\ \hline
AlSalamah et al. \cite{salamah2018towards} & 1,000            & Characters                 &  Arabic   & $\checkmark$           \\ \hline
Koudja et al. \cite{kaoudja2019efficient}  & 1,685          & Sentences                    &  Arabic  & $\checkmark$            \\ \hline
KERTAS \cite{adam2018kertas}               &  2,000             & Historical Documents                     & Arabic     & $\checkmark$         \\ \hline
Khayyat et al. \cite{khayyat2020deep}      & 2,653             & Historical Documents                   &  Arabic &            \\ \hline
ADAB \cite{margner2009icdar}               & 32,492            & Words &  Arabic  & $\checkmark$             \\ \hline
Slimane et al. \cite{slimane2009new}       &  113,284          & Words                    & Arabic   & $\checkmark$           \\ \hline
Bataineh et al. \cite{bataineh2013arabic}  & 700              & Sentences                    &  Arabic/Jawi &         \\ \hline
UPTI \cite{sabbour2013segmentation}        &  10,000         & Sentences          & Arabic/Urdu  & $\checkmark$             \\ \hline
MAYASTROUN  \cite{njah2012mayastroun}           & 67,825           & Sentences                  &  Arabic/English & $\checkmark$   \\ \hline
\end{tabular}
\caption{Summary of datasets in the literature. }
\label{tab:lit}

\end{table}

\section{Annotation, Representation and Exploration}

\subsection{Annotation}
\begin{figure}[htp!]

\centering
\includegraphics[width=0.5\textwidth]{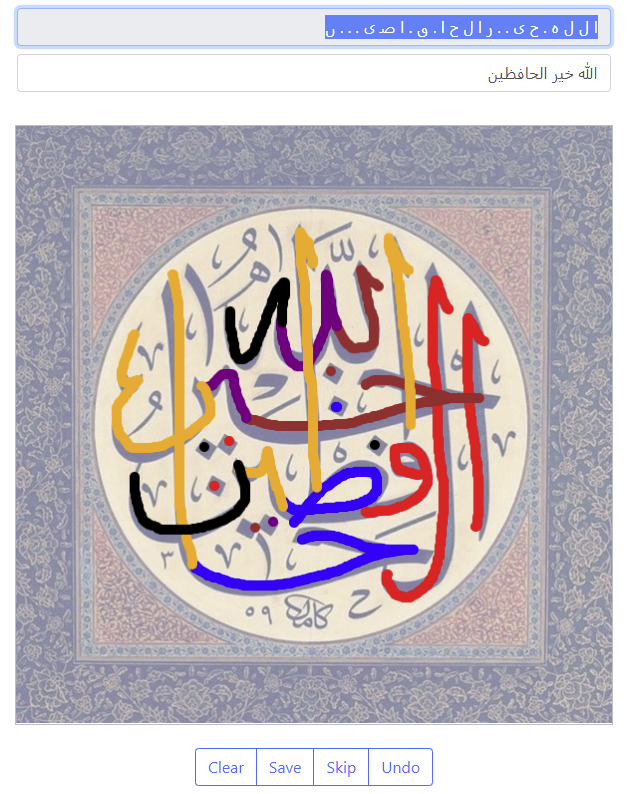}
\caption{The annotation process.}
\label{fig:annot}
\end{figure}

The initial phase started by scraping the internet of Arabic calligraphy. We collected a wide variety of calligraphy like Diwani, Thuluth, Kufi, Farsi and many others. The annotation was implemented by tracking the calligraphy images using a digital pen and tablet by four different annotators. In Figure \ref{fig:annot}, we present an example of the annotation process. The main canvas in the center contains the image to be annotated. The different colors used in the annotation represent different strokes. At the bottom, we have different buttons to simplify the annotation process like undoing a certain stroke in addition to being able to clear or skip the current drawing. There are mainly two input fields at the top of the annotated image. In the second input, we see the annotation text which is editable so the annotator can write new annotations or fix any misspelled data instances. In the top input field, we split up each letter into simple components where each component represents a stroke. A stroke is defined as a single writing procedure where the writer doesn't lift their hand.  Hence, a lot of letters will be split into different components depending on how they are written as shown in Table \ref{tab:decom}. For example, letters that contain dots or nekat \<نقاط> in Arabic are split into the basic letter without the dot and the dot as a different stroke. We also differentiate between dots that appear above or below the letter. If the dot appears above the letter we write it before the letter itself, otherwise it comes after. This also extends to other special characters like Hamza \<ء> which can appear below or above a certain letter. The point behind encoding that in an explicit format is to guide the annotator how to write each letter and whether to ignore certain characters like special characters. In our annotation process diacritics and other characters that are used for decoration are ignored. We only include Maddah $\sim$ which usually happens with Alef \<ا> . In total we have 37 unique letters in our vocabulary. 

\begin{table}[htp!]
\centering
\begin{tabular}{l|l||l|l}
\hline 
\textbf{Letter} & \textbf{Decomposition}          & \textbf{Letter} & \textbf{Decomposition}           \\ \hline
\<أ>             & \<ا ء> & \<آ >    & $\sim$ \<ا>   \\ \hline
\<إ>     & \<ء ا>            & \<ب>     & \< . ٮ >                  \\ \hline
 \<ت>     &  \<ٮ . .  >              &   \<ث> &  \<ٮ . .  . >                        \\ \hline
\<خ>     & \<ح .>   & \<ج>     & \<. ح>               \\ \hline
\<ذ>     & \<د .>          & \<ز>     & \<ر .>           \\ \hline
\<ش>     & \<س . .  . >               &  \<ض>    & \<ص .>          \\ \hline
\<ط>             & \<صـ ا>  &  \<ظ>             & \<صـ ا .> \\   \hline
\<غ>     & \<ع .>                     & \<ف>     & \<ٯ .>  \\ \hline
\<ق>     & \<ٯ . .>       & \<ك>             & \<ل ء >   \\ \hline
 \<ن>     & \<ٮ  .>  &     \<ؤ>             & \<و ء >  \\ \hline 
 \<ة>     &  \<ه . .  >    &  \<ئ>             & \< ى ء >   \\ \hline
\end{tabular}
\caption{Decomposition of letters into basic components.}
\label{tab:decom}
\end{table}

\subsection{Representation}
\begin{figure}
    \centering
    \includegraphics[width=0.7\textwidth]{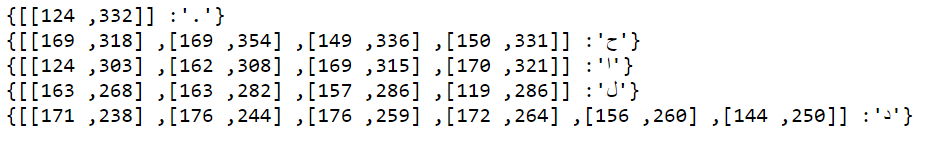}

\caption{Raw data for the word \<خالد> .}
    \label{fig:raw_data}
\end{figure}

\begin{figure}[htp!]

\centering
\includegraphics[width=0.6\textwidth]{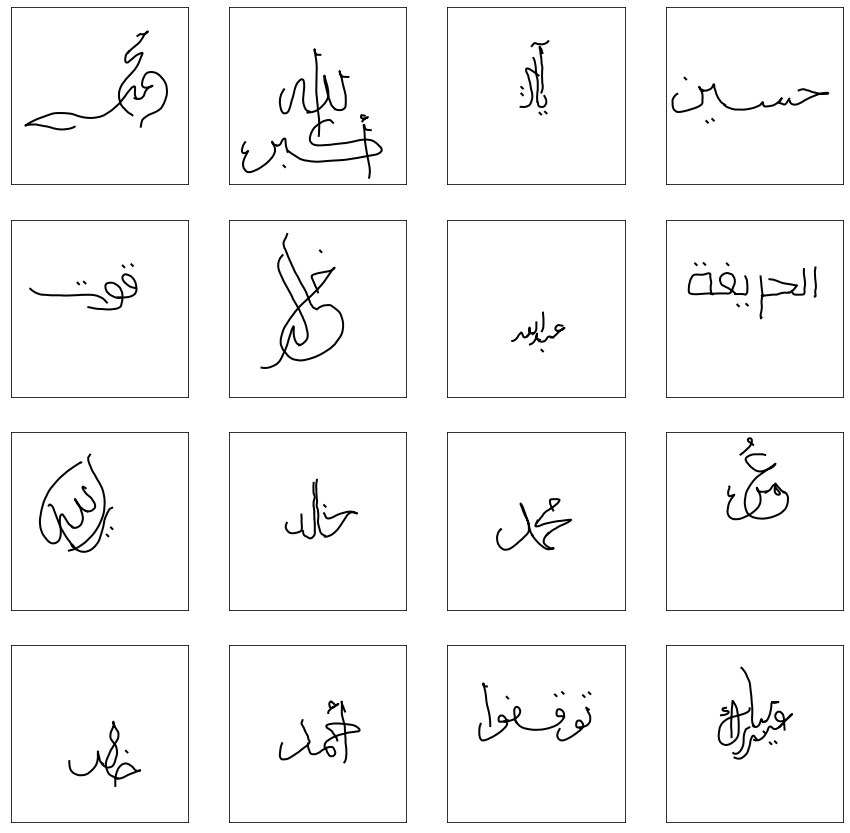}
\caption{Examples of the visualizing the json files. }
\label{fig:sketches}
\end{figure}

We tried to come up with a consistent representation such that researchers can easily reuse the dataset. In this manner, the annotation process is carefully designed. When the annotator finishes a certain stroke, it is recorded as a list of $(x, y)$ coordinates representing where the pen traveled while drawing the stroke. We define a sketch as a list of strokes and each stroke consists of the annotation i.e what kind of letter was written in addition to the $(x,y)$ components. Each sketch is saved as json file and can be easily converted to an image by drawing the coordinates. In order to avoid the problem of having images with different dimensions, we fix the maximum size of each dimension i.e width or height to 600 pixels. In order to keep the aspect ratio of the images and the strokes, the larger dimension was scaled to 600 pixels and the smaller dimension was multiplied by the same factor. As an example, if an image had a size (400, 200) then we rescale it to the size (600, 300). In Figure \ref{fig:sketches}, we show 16 images by redrawing the content of the json files. The dataset is publicly available at GitHub \footnote{\url{https://github.com/ARBML/Calliar}}. The compressed zip format of the dataset has size of around 52 MB. We also create a compressed format of size 8.6 MB which has the same format as QuickDraw \cite{ha2017neural}. The compressed format doesn't have any stroke level annotations and  the Ramer-Douglas-Peucker (RDP) algorithm is applied to reduce the number of points \cite{douglas1973algorithms}. In Figure \ref{fig:raw_data}, we visualize the raw data of a given word \<خالد>
. Note that for visual purposes we apply the RDP algorithm.  The raw data contains a list of dictionaries where each dictionary represents the stroke as text and a list of points. 

\subsection{Exploration}
\begin{figure}

     \begin{subfigure}[b]{0.40\textwidth}
         \includegraphics[width=\textwidth]{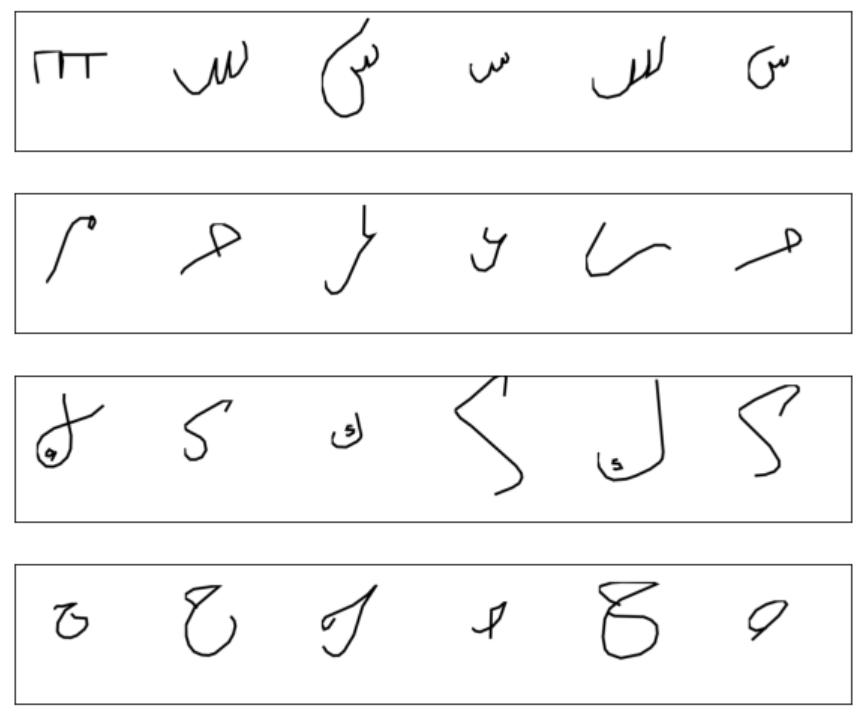}
         \caption{Sample character variations.}
         \label{fig:char_variations}
     \end{subfigure}
     \hfill
     \begin{subfigure}[b]{0.40\textwidth}
\includegraphics[width=0.97\textwidth]{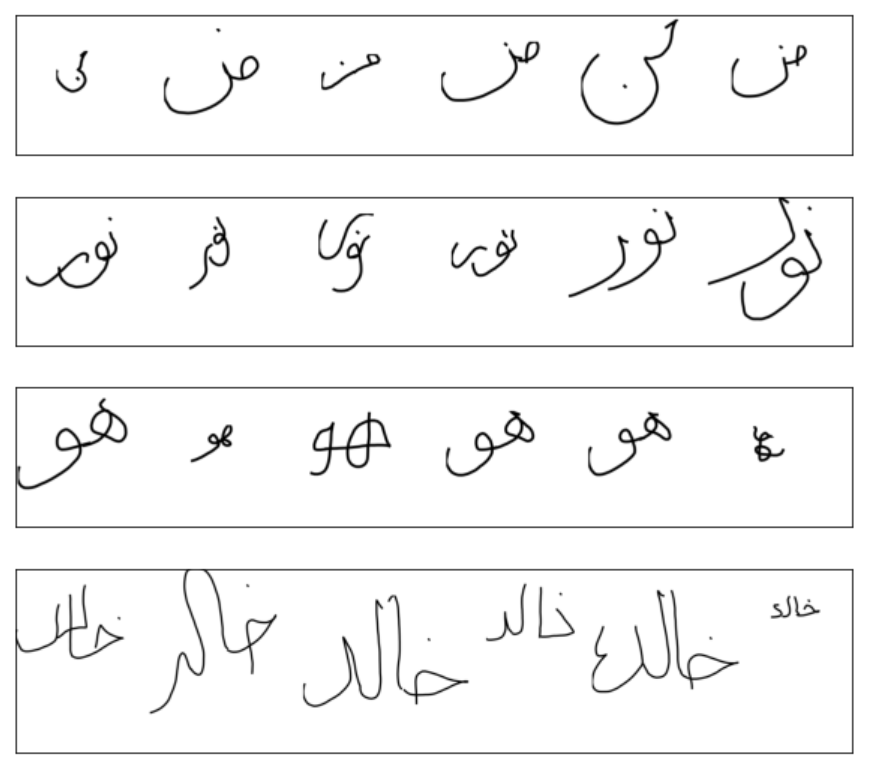}
         \caption{Sample word variations.}
         \label{fig:word_variations}
     \end{subfigure}
    \caption{Different variations  of drawing characters and words.  }
    \label{fig:variations}
\end{figure}

Data visualization and exploration is a necessary technique to assess the quality of the collected dataset. As we saw from Figure \ref{fig:sketches}, there is some kind of variations in writing the same letter. In Arabic script writing, letters are written connected together which causes a variation between a letter in beginning, middle or at the end. Moreover, various calligraphy styles draw characters differently which might cause a single letter to be confused with another. This is much more apparent when we try visualizing individual characters in our datasets. In Figure \ref{fig:char_variations}, we visualize four different letters sampled randomly from our dataset. For example, the letter \<س> , in the first row is written in different variations. The first image of the character is interesting because the character is drawn upside down. This type of drawing mostly happens in Kufic calligraphy. The letter \<م> is also interesting because it can be written with or without a loop depending on the style of calligraphy.  The letter \<ك> is unique because it can appear in either single stroke or double stroke depending in its position on a given word. Similarly in Figure \ref{fig:word_variations}, we see the results of visualizing word level representations. This is important because this proves that our dataset can be used to extract word-level features which can be utilized for word level recognition.  The most interesting one is the writing of the word \<هو> which can be written in many variations as seen in the third row. One thing that makes this dataset complex is the freedom of writing in different angles like in the last column of the third row where the word is written vertically. 

\begin{figure}[htp!]

\centering
\includegraphics[width=0.6\textwidth]{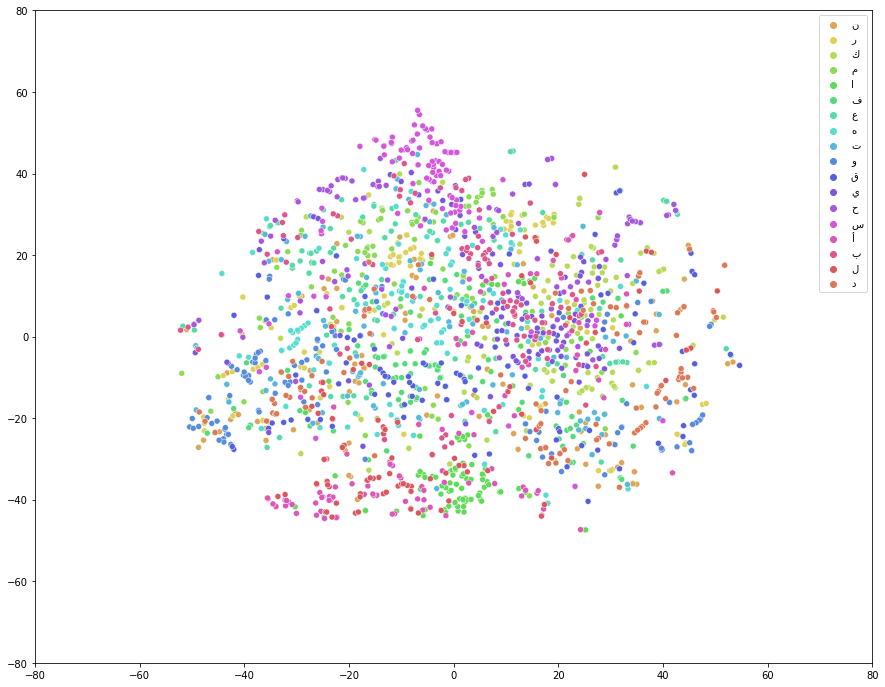}

\caption{Visualizing top characters using t-SNE algorithm. }
\label{fig:tsne}
\end{figure}

In Table \ref{tab:levels}, we summarize the annotation levels of the dataset. Each level is annotated and can be extracted easily from the dataset with its corresponding annotation. The first level represents an annotation for the full drawn calligraphy, while the second level represents an annotation for the word level. The third level is an annotation on the character level and finally, the fourth level is an annotation on the stroke level.  Such granularity of representation allows our dataset to be used in multiple tasks ranging from stroke and character to word and sentence level prediction. In Figure \ref{fig:tsne}, we visualize a subset of the characters in the dataset using the t-SNE algorithm \cite{van2008visualizing} for data projection. We extract 100 images from 18 characters and crop and resize to around $64 \times 64$ then apply the algorithm to reduce the dimension to 2D space. We see from the projections that the algorithm was able to cluster some letters like \<س> and \<ح>. We also notice that the clusters for \<ا> and \<ل> are close which is expected because they usually have similar sketches especially in connected letters.  

{
\begin{table}[htp!]

\centering
\begin{tabular}{Sl|Sc}
\hline
Level 1: full sketch  &
\includegraphics[width=0.15\textwidth]{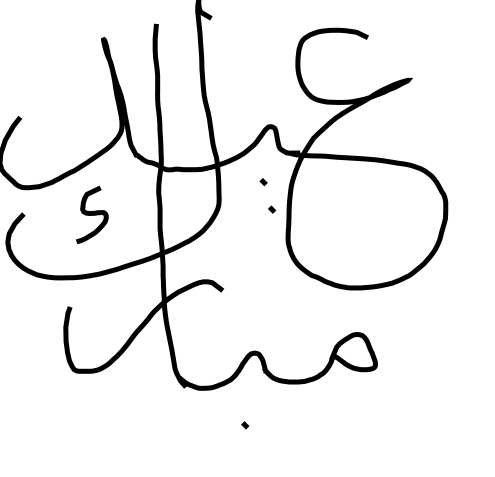} 
   \\ \hline
Level 2: word sketches & \includegraphics[width=0.27\textwidth]{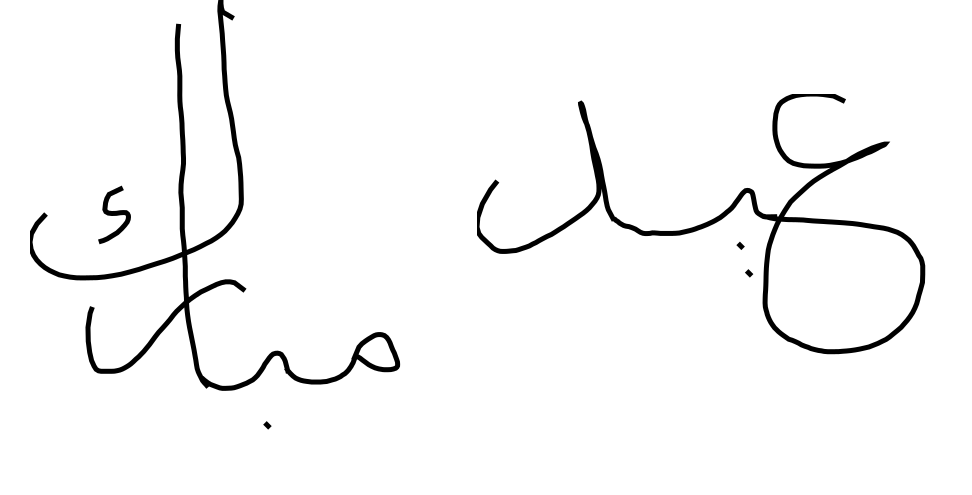}
   \\ \hline
Level 3: character sketches  & \includegraphics[width=0.5\textwidth]{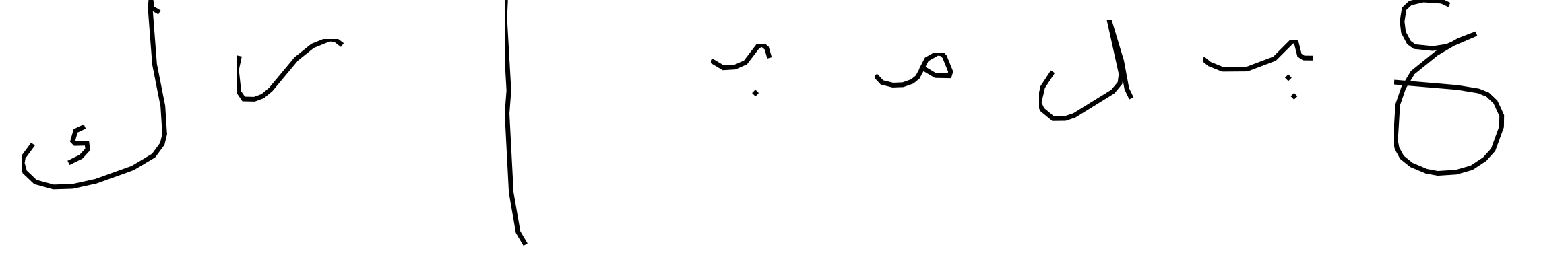}
   \\ \hline
Level 4: stroke sketches  & \includegraphics[width=0.6\textwidth]{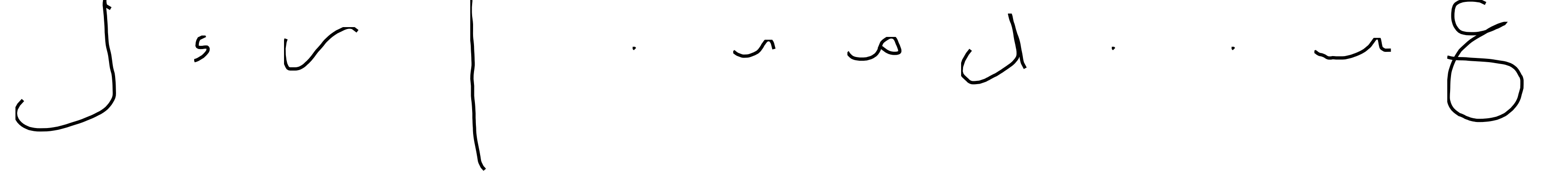}
   \\ \hline
\end{tabular}
\caption{Levels of drawing a sample image containing the text \<عيد مبارك>.}
\label{tab:levels}
\end{table}
}

\begin{figure}[htp!]
\centering
\includegraphics[width=0.6\textwidth]{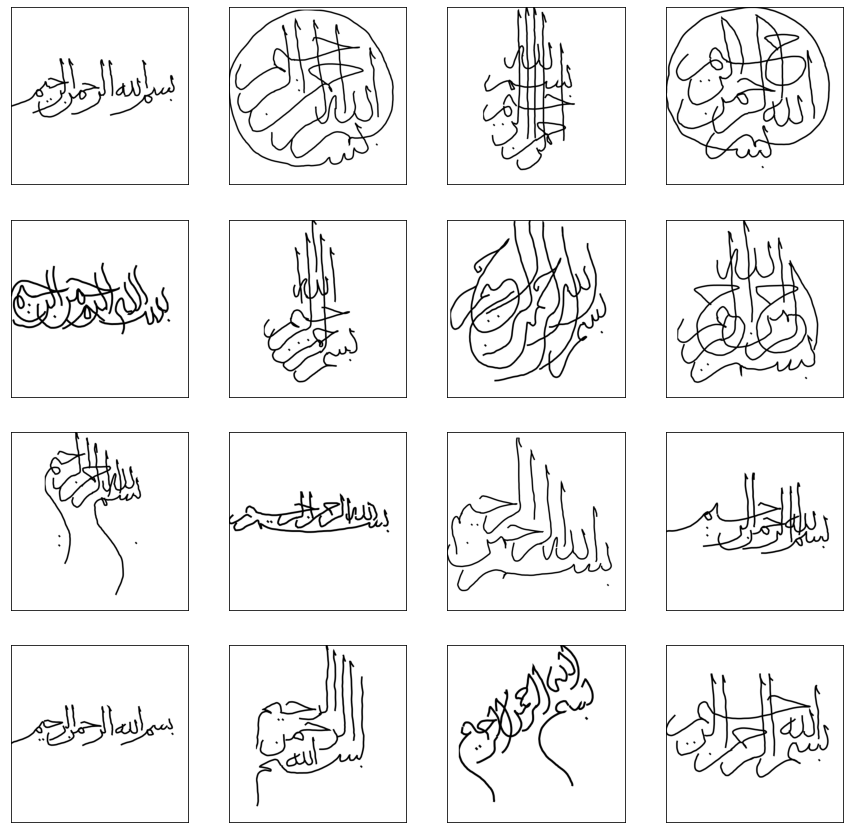}

\caption{Visualizing 16x16 grid of the sentence \<بسم الله الرحمن الرحيم>. }
\label{fig:sent_var}
\end{figure}

In Figure \ref{fig:sent_var} we see the variations of drawing the sentence \<بسم  الله الرحمن الرحيم>. Which is one of the most frequent sentences that writers like to write. 
We see the complexity in style of drawing the same sentence which adds an extra dimension of artistic view for Arabic calligraphy 
For a given calligraphic style we can sketch a sentence in numerous styles. One style that seems prominent is the alignments of all the Alef characters \<ا> which adds some kind of symmetry to the drawing as we see in many examples in the figure.

\section{Dataset Statistics}
\begin{figure}

    \centering
    \includegraphics[width=0.7\textwidth]{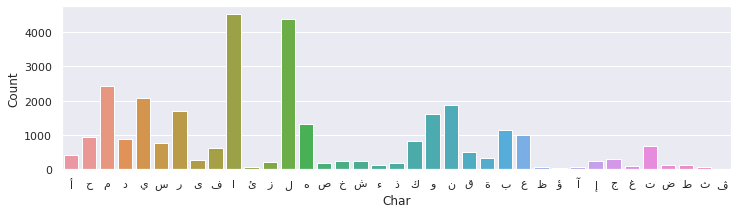}
        \caption{Character histograms across all the dataset.}
    \label{fig:char_counts}

\end{figure}

In Table \ref{tab:stats}, we summarize the statistics about the collected dataset. We split the dataset randomly into training, validation and testing and we show the number of strokes, characters, words and sentences in each split.  In Figure \ref{fig:char_counts}, we show the histogram of all characters across all the datasets. Mainly, we see the characters with high frequency are those which are mostly used in Arabic writing like \<ا , ل , م > . Note that most of the characters with low frequency are composite characters like 
\<
ؤ ، ئ ، آ 
> . Along with some characters that are rarely used in Arabic writing like \<
ظ ، غ ، ڤ
> . Note that, many characters might happen most frequently because of the imbalance in the representations in some calligraphic images. For example, the sentence \<
بسم الله الرحمن الرحيم
> happen a lot in the writing for calligraphy. The reason of including the same sentence multiple times is to increase the diversity of drawing the same sentence as we see in Figure \ref{fig:sent_var}.

\begin{table}[htp!]
\begin{center}

\begin{tabular}{l|c|c|c|c}
\hline
 & \textbf{Number of Samples} & \textbf{Number of Words} & \textbf{Number of Chars} & \textbf{Number of Strokes}  \\ \hline \hline
\textbf{Train} & 2,000 & 6,065 & 24,722 & 36,561 \\ \hline
\textbf{Valid} & 250 & 738 & 2,946 & 4,410 \\ \hline
\textbf{Test} & 250 & 753 & 3,052 & 4,601 \\ \hline
\hline
\textbf{Total} & 2,500 & 7,556 & 30,720 & 45,572 \\ \hline
\end{tabular}
\end{center}
\caption{Data statistics. }
\label{tab:stats}
\end{table}

\section{Research Impact}
\begin{figure}[htp!]
\begin{subfigure}[b]{0.5\textwidth}\centering\includegraphics[width=0.3\textwidth]{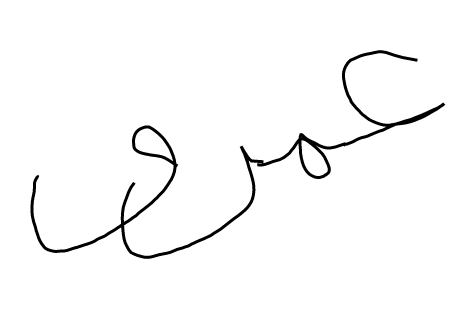}\caption{Naskh}\label{fig:taba}\end{subfigure}
     \hfill
\begin{subfigure}[b]{0.5\textwidth}\centering\includegraphics[width=0.5\textwidth]{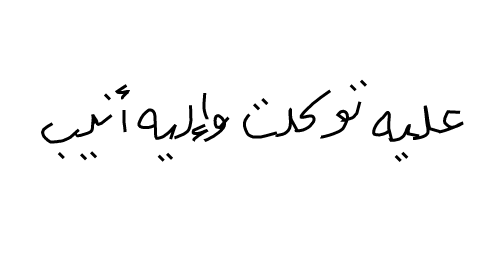}\caption{Rekaa }\label{fig:tabb}\end{subfigure}

\begin{subfigure}{0.5\textwidth}\centering\includegraphics[width=0.4\columnwidth]{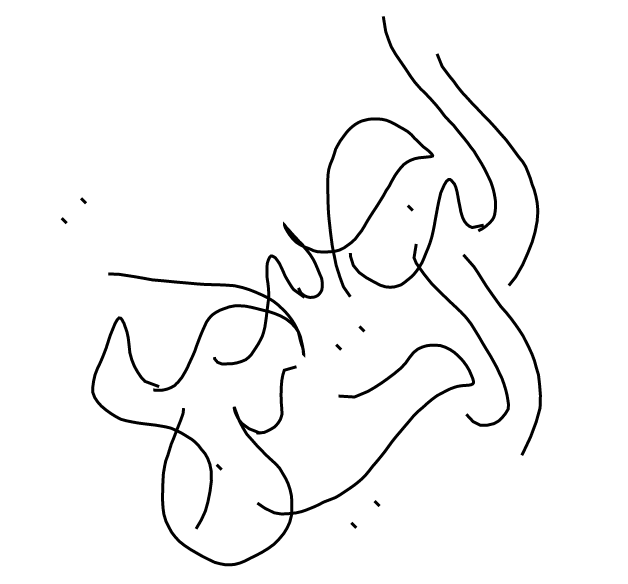}\caption{Wisam}\end{subfigure}
\begin{subfigure}{0.5\textwidth}\centering\includegraphics[width=0.4\columnwidth]{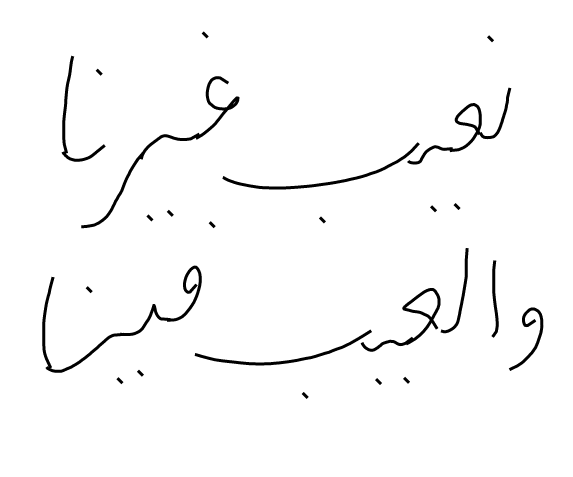}\caption{Farisi}\end{subfigure}

\begin{subfigure}{0.5\textwidth}\centering\includegraphics[width=0.4\columnwidth]{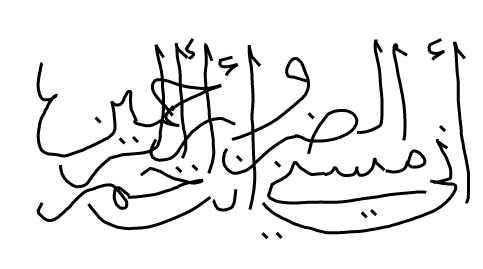}\caption{Thuluth}\end{subfigure}
\begin{subfigure}{0.5\textwidth}\centering\includegraphics[width=0.4\columnwidth]{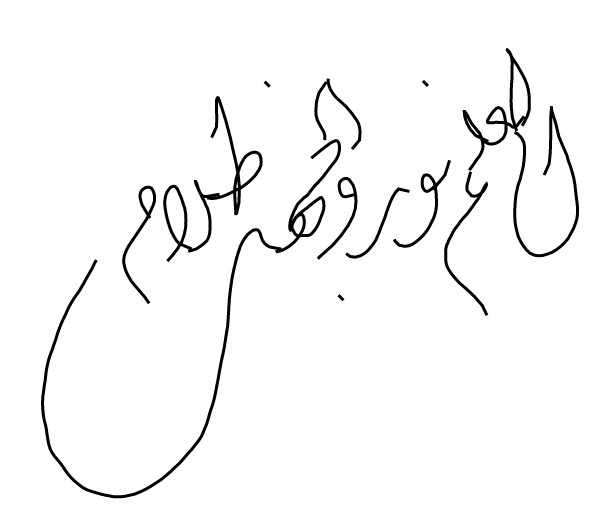}\caption{Diwani}\end{subfigure}
\caption{Different styles of calligraphy.}
\label{fig:styles}
\end{figure}

The stroke-level annotations of the dataset allow it to be used in a wide range of research like text classification, optical character recognition (OCR), and sketch generation. The text classification can range from character level classification to complete sentence classification and prediction. Arabic OCR is a difficult task because it contains mostly letters that are written connected together. Our dataset makes annotations on the stroke and character level in many variations and styles which allows training efficient models for such tasks. Such examples of these variations could be found in Figure \ref{fig:styles}. The artistic aspect is also important for this dataset. In the literature, there is no dataset that captures such level of details for Arabic calligraphy in an online format. Moreover, for OCR for instance, we need to preprocess the dataset by segmenting individual letters which is not necessary in our dataset. 

In addition to the above-mentioned examples, the dataset could be used for sketch generation as a creative application. Most of the research in the literature that deals with sketch drawing and text-to-image, consider English like GANwriting \cite{kang2020ganwriting}, Scrabble-GANs \cite{fogel2020scrabblegan}, DF-GANs \cite{tao2020df}, BézierSketch \cite{das2020beziersketch}, sketchRNN \cite{ha2017neural} and DoodlerGAN \cite{ge2020creative}. This field of research is very important because it combines multiple modalities ranging from natural language processing (NLP) , generative adverserial networks (GANs) and  creative applications. In the literature, there are hundreds of papers published in each year for each of those fields but due to the lack of proper datasets, there are no real advancements in making systems that deal with all of them especially for Arabic. We believe that this dataset can fill this gap. 

\section{Social and Historical Impact} 
Arabic calligraphy has a long history of being used in different communities in the Arabic world. Each community embraced a specific style of calligraphy which resulted in many calligraphic styles across the Arabic world. For instance, the Moroccan calligraphy style "Al Maghribi" might contain up to five different types of calligraphy \cite{chaker2011creation}, which is considered an important part of the Moroccan heritage and culture. Without a proper way of making such calligraphy digital by encoding the specific drawing of each letter as written by original writers, such calligraphy might disappear in the future. Moreover, encoding different calligraphic styles can help in deciphering some of the old calligraphic styles which are used in historical manuscripts.  

In the current digital world, a lot of companies are using calligraphy as a way of delivering information and capturing the attention of possible customers. Looking at advertisements on social media and on billboards, there is an increase interest in making use of different calligraphic styles. Moreover, during holidays and vacations, people use calligraphic written posts to congratulate each other during different holy events. We believe that our dataset can be used for automating the process of generating different calligraphic styles.  

\section{Conclusion}
In this paper, we introduced Calliar which is an online dataset for Arabic calligraphy that contains 2,500 sentences and more than 40,000 strokes. The dataset allows capturing  calligraphy in multiple levels ranging from stroke, character, word to sentence level representation. The granularity level allows for using the dataset in multiple tasks like classification on the character or the word level. In addition to that, we can use the dataset for calligraphy generation and character recognition. The dataset consists of a wide range of calligraphic styles like Diwani, Thuluth, Farisi, etc. The different styles make the dataset unique because of the complexity of drawing letters in each of the styles.  As a future work, we are planning to release new versions which contain multiple variations in addition to increasing the size of the dataset massively. This is important because training deep networks usually require thousands of samples, and as the dataset size increases, the quality of the results increases. Moreover, we would like to balance the number of samples of each character to increase the probability of some rare characters like \<ث> and \<ض>.    

\bibliographystyle{unsrt}  
\bibliography{references}  


\end{document}